\documentclass{article}

\usepackage{microtype}
\usepackage{graphicx}
\usepackage{subfigure}
\usepackage{booktabs} 
\usepackage{tcolorbox}
\usepackage{hyperref}
\usepackage{float}
\usepackage{bm}

\usepackage[accepted]{icml2025}

\usepackage{amsmath}
\usepackage{amssymb}
\usepackage{mathtools}
\usepackage{amsthm}

\usepackage[capitalize,noabbrev]{cleveref}

\theoremstyle{plain}

\theoremstyle{definition}

\theoremstyle{remark}

\usepackage[textsize=tiny]{todonotes}
\newcommand{\logdet}{\log\det}
\newcommand{\norm}[1]{\left\| #1 \right\|}
\icmltitlerunning{PRISM: Deriving Transformer as a Signal-Noise Decomposition Operator via Maximum Coding Rate Reduction}

\begin{document}

\twocolumn[
\icmltitle{PRISM: Deriving a White-Box Transformer as a Signal-Noise Decomposition Operator via Maximum Coding Rate Reduction}



\icmlsetsymbol{equal}{*}

\begin{icmlauthorlist}
\icmlauthor{Dongchen Huang}{IOP}

\end{icmlauthorlist}

\icmlaffiliation{IOP}{Institute of Physics, Chinese Academy of Sciences}

\icmlcorrespondingauthor{Dongchen Huang}{huangdongchen@iphy.ac.cn}

\icmlkeywords{Transformer, White-box Deep Learning}

\vskip 0.3in
]


\printAffiliationsAndNotice{} 

\begin{abstract}
Deep learning models, particularly Transformers, are often criticized as "black boxes" and lack interpretability. We propose Prism, a white-box attention-based architecture derived from the principles of Maximizing Coding Rate Reduction ($\text{MCR}^2$). By modeling the attention mechanism as a gradient ascent process on a distinct signal-noise manifold, we introduce a specific irrational frequency separation ($\pi$-RoPE) to enforce incoherence between signal (semantic) and noise (syntactic) subspaces. We show empirical evidence that these geometric inductive biases can induce  unsupervised functional disentanglement alone. Prism spontaneously specializes its attention heads into spectrally distinct regimes: low-frequency heads capturing long-range causal dependencies (signal) and high-frequency heads handling local syntactic constraints and structural artifacts. To provide a theoretical grounding for these spectral phenomena, we draw an analogy between attention mechanism and a Hamiltonian dynamical system and identify that the standard geometric progression of Rotary Positional Embeddings (RoPE) induces dense resonance networks (Arnold Tongues), leading to feature rank collapse. Empirical validation on 124M-parameter models trained on OpenWebText demonstrates that Prism spontaneously isolates the Attention Sink pathology and maintains isentropic information flow across layers. Further, we suggest a physics-informed plug-and-play intervention KAM-RoPE for large language models (LLMs).  Our results suggest that interpretability and performance can be unified through principled geometric construction, offering a theoretically grounded alternative to heuristic architectural modifications
\end{abstract}

\section{Introduction}
\label{Introduction}

The Transformer architecture \cite{vaswani2017attention} has established itself as the fundamental building block of modern artificial intelligence, exhibiting a remarkable power-law relationship between parameter scale and performance \cite{kaplan2020scaling}. However, despite their empirical success, these models operate largely as black boxes. One limitation is the architectural conflation between semantic reasoning (the ability to manipulate abstract concepts over long-range contexts) and syntactic memorization (the rote statistical completion of local patterns). While recent mechanistic interpretability efforts have identified specific circuits such as induction heads \cite{olsson2022incontext}, the standard attention mechanism lacks the physical inductive bias required to  separate these distinct functional regimes by design. 

As these models scale, distinct phenomenological anomalies have emerged. Notably, standard causal Transformers tend to "dump" excessive probability mass onto the initial token, a phenomenon called "Attention Sink" \cite{Xiao2024}, and frequently collapse into degenerative repetition loops under long-context generation. We propose that these are not merely engineering artifacts, but symptoms of a feature rank collapse: a structural failure where the attention mechanism, trapped by spectral resonances, loses the capacity to distinguish informative signals from high-frequency noise in deep layers. These limitations may often result in brittle reasoning, hallucination driven by overfitting to noise and data inefficiency.

To enhance interpretability, we turn to first-principles approaches that view deep learning as data compression. The principle of Maximal Coding Rate Reduction ($MCR^2$) \cite{Chan2022ReduNet} shows that the fundamental goal of representation learning is to expand the volume of the dataset globally while compressing the volume of similar classes locally. Building on this theoretical foundation, the CRATE (Coding-RATE transformer) framework \cite{Yu2024White-Box} establishes that the self-attention mechanism is mathematically isomorphic to a step of gradient ascent optimizing the rate reduction. This perspective transforms the attention layer mechanism from an empirical heuristic into a derived operator for signal denoising on low-dimensional manifolds, providing the theoretical foundation for our architectural interventions.

However, gaps between white-box neural networks and black-box neural networks still exist. One critical gap lies in data modeling; current mathematically interpretable white-box models (e.g., unfolded optimization networks) typically assume clean data with Gaussian noise \cite{Chan2022ReduNet}, whereas the noise distributions of real-world data are highly non-Gaussian and structural. This discrepancy leads to a conflict in the optimization trajectory. We posit that in the context of global reasoning, the definition of ``noise'' should be relative, structured and scale-dependent.  While local syntactic constraints (e.g., grammar, n-gram statistics) are essential for surface-level linguistic coherence, they manifest as high-frequency interference against the low-frequency global dependencies (semantic signals) required for long-range reasoning. In standard Transformers with limited capacity, the gradient flows of these two regimes undergo spectral entanglement. The high-energy local syntactic artifacts often overshadow the sparse global semantic signals, hindering the optimization process and leading to feature rank collapse.

To mitigate this, we propose Prism, a physics-informed transformer architecture that introduces two specific geometric constraints to the standard Transformer block based on explicit additive noise. First, we impose an overcomplete dictionary \cite{candes2006robust} to expand the representational phase space, allowing signal and noise to separate in higher dimensions without competing for the same limited basis vectors.  Second, we implement frequency separation via $\pi$-scaled Rotary Positional Embeddings ($\pi$-RoPE) \cite{su2024roformer} to encourage the incoherence between signal and noise based on resonance. By scaling the rotational bases of the overcomplete subspaces by $\pi$ and $1/\pi$ respectively, we propose that this choice effectively mitigates resonance and keeps the reasoning and memorization dynamics on spectrally disjoint manifolds.

We empirically validate the Prism architecture under controlled constraints using a compact model with 50M parameters trained on the TinyStories benchmark \cite{eldan2023tinystories}. Despite the restricted scale, our experiments suggest that Prism reveals a phenomenon of unsupervised functional emergence and good training stability. We observe that the attention heads spontaneously organize into spectrally distinct regimes: heads driven by the low-frequency basis specialize in tracking long-range correlations (e.g. semantic signal), while those driven by the high-frequency basis tend to capture local correlations (e.g. syntactic structure). These results provide a proof-of-concept that disentangled reasoning capabilities can be induced via principled geometric construction rather than solely through massive parameter scaling.

For real-world datasets, we validate Prism by scaling it to 124M parameters on the OpenWebText corpus. Despite utilizing standard training recipes, Prism exhibits a distinct thermodynamic signature compared to the GPT-2 baseline. Most notably, we observe the absence of the Attention Sink phenomenon and the preservation of high attention entropy (Isentropic Flow) throughout the network depth, and this can be interpreted by mapping the attention mechanism to a classical Hamiltonian system and studying its perturbation. 

Our results suggest that by enforcing rigorous geometric and physical constraints—specifically the orthogonality between signal and noise, we can construct "white-box" Transformers that are not only interpretable by design but also dynamically robust. Preliminary theoretical analysis suggests a path to physics-informed interventions (KAM-RoPE) for existing LLMs.

\section{Theoretical Framework}

\subsection{Rate Reduction as the Objective}
We begin with the assumption that a useful signal $\mathbf{Z}$ can be understood as the combination of compressible useful $\mathbf{Z}_0$ and the noise $\mathbf{E}$, which yields:
\begin{equation}
    \begin{split}
        \mathbf{Z} = \mathbf{Z}_0 + \mathbf{E},
    \end{split}
\end{equation} 
and the learning objective is to extract $\mathbf{Z}_0$ from $\mathbf{Z}$. To achieve this, we follow the maximum coding rate reduction framework. Let $\mathbf{Z} \in \mathbb{R}^{N \times d}$ denote representation (e.g. input tokens), and $\mathbf{U}_s,\mathbf{U}_n $ be the subspace with respect to the effective signal $\mathbf{Z}_0$ and noise $\mathbf{E}$ respectively. We assume the signal and noise reside in distinct subspaces, and these subspaces $\mathbf{U}_s$ and $\mathbf{U}_n$ are sufficiently incoherent (e.g., $\norm{\mathbf{U}_s^\top \mathbf{U}_n} \approx 0$). We can approximate the noise $\mathbf{E} = \mathbf{Z}\mathbf{U}_n $. Thus, the effective signal $Z_0$ is obtained by maximizing the coding rate difference $\Delta R(\mathbf{Z})$:

\begin{equation}
    \mathbf{Z}_0 = \max_{\mathbf{Z}} \Delta R(\mathbf{Z}) = R(\mathbf{Z}|\mathbf{U}_s) - \lambda R(\mathbf{Z}|\mathbf{U}_n),
    \label{eq:rate_reduction}
\end{equation}
where $R(\mathbf{Z})$ represents the coding rate of the signal projected onto local subspaces expanded by $\mathbf{U}_s$, and $R(\mathbf{Z}|\mathbf{U}_n)$ represents the coding rate of the noise obtained by projecting the noisy signal onto the local subspaces $\mathbf{U}_n$, and $\lambda>0$ is the parameter controlling the decomposition between signal and noise. While the closed-form expression of the coding rate is originally derived under the Gaussian assumption, we adopt it here as a robust geometric proxy for measuring the volume of the feature manifold. The coding rate of the matrix $\mathbf{Z}$ is defined as:

\begin{equation}
    R(\mathbf{Z}) = \frac{1}{2} \log \det \left( \mathbf{I} + \frac{d}{N \epsilon^2} \mathbf{Z}\mathbf{Z}^\top \right),
\end{equation}
where $\epsilon$ is the distortion of encoding.

If there are $K$ subspaces $\mathbf{U}=[\mathbf{U}_1,\dots,\mathbf{U}_K]$, the coding rate of $\mathbf{Z}$ is the sum of the coding rates of all subspace projections, and yields
\begin{equation}
R(\mathbf{Z} \mathbf{U}_k)  = \frac{1}{2}\logdet\left(\mathbf{I} + \beta (\mathbf{Z} \mathbf{U}_k) (\mathbf{Z} \mathbf{U}_k)^\top\right),
\end{equation}
where $\beta = \frac{p}{N\epsilon^2}$ and $p$ is the dimension of subspaces and $k=1,\dots,K.$

\subsection{Prism as Gradient Ascent of Coding Rate Reduction}
\begin{figure*}[t]
  \centering
  \includegraphics[width=0.7\textwidth]{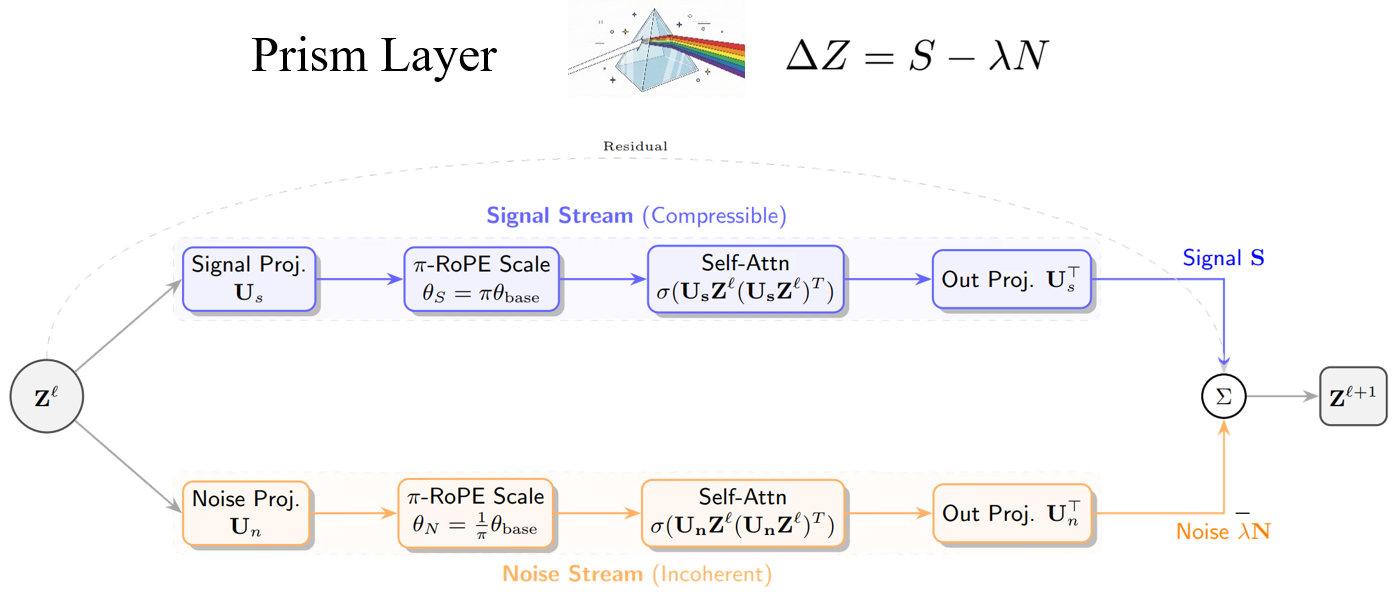}
  \caption{\textbf{The PRISM White-Box Architecture.} 
    \textbf{(Top)} The input latent state $\mathbf{Z}^\ell$ is projected into an overcomplete feature space via dictionary $\mathbf{U}=[\mathbf{U}_s,\mathbf{U}_n]$. The data flow in Prism bifurcates into \textbf{Signal Stream} (Blue) and \textbf{Noise Stream} (Orange). 
    \textbf{$\pi$-RoPE}: the signal stream applies low-frequency rotary embeddings scaled by $\pi$ ($\theta_S = \pi\theta_{\mathrm{base}}$) to represent long-range semantic structures, while the noise stream applies high-frequency embeddings ($\theta_N = \frac{1}{\pi}\theta_{\mathrm{base}}$) to capture short-range syntactic artifacts. The 'noise stream' is explicitly designed to capture and isolate syntactic artifacts or short-range information in NLP tasks.
    The layer output is computed via a differential operator, $\mathbf{Z}^{\ell+1} \leftarrow S - \lambda N$, where $\lambda$ is an annealing coefficient that dynamically suppresses the noise subspace, encouraging the model to learn long-term correlation (e.g. causal logic) during training.}
    \label{fig:prism_architecture}
  \label{fig:prism}
\end{figure*}

To optimize the representations $\mathbf{Z}$ and remove the noise layer-by-layer, we apply a single step of gradient ascent on the objective $\Delta R(\mathbf{Z})$ with step size $\eta$. This yields the dynamic update rule similar to the Transformer block:

\begin{equation}
    \mathbf{Z}^{\ell+1} = \mathbf{Z}^{\ell} + \eta \frac{\partial \Delta R(\mathbf{Z}^\ell)}{\partial \mathbf{Z}^\ell}.
\end{equation}
The Prism block is obtained by deriving the gradients for the log-determinant terms. Following \cite{Yu2024White-Box}, the derivative of the log-determinant term yields the softmax structure. We recover the structure of the one-head self-attention mechanism under the white-box constraint $\mathbf{Q}=\mathbf{K}=\mathbf{V}=\mathbf{U}$ and the output matrix is set directly into $\mathbf{W}_O = \mathbf{U}^\top$:

\begin{equation}
\begin{split}
\mathbf{Z}^{\ell+1} \approx &\mathbf{Z}^{\ell} + \eta \underbrace{\text{softmax}\left( \frac{(\mathbf{Z}^\ell \mathbf{U}_s) (\mathbf{Z}^\ell \mathbf{U}_s)^\top}{\sqrt{d_k}} \right)}_{\text{Attention Matrix}} (\mathbf{Z}^\ell \mathbf{U}_s) \mathbf{U}^\top_s \\
    &- \lambda \eta \underbrace{\text{softmax}\left( \frac{(\mathbf{Z}^\ell \mathbf{U}_n) (\mathbf{Z}^\ell \mathbf{U}_n)^\top}{\sqrt{d_k}} \right)}_{\text{Attention Matrix}} (\mathbf{Z}^\ell \mathbf{U}_n) \mathbf{U}_n^\top.
\end{split}
\label{eq:gradient_ascent}
\end{equation}

Here, the term $\mathbf{Z}^\ell \mathbf{U}_s$ represents the projection onto the subspace $\mathbf{U}_s$, effectively separating the noise if the noise is incoherent with the useful signal, and $d_k$ is the dimension of the space $\mathbf{U}_s$ or $\mathbf{U}_n$.

An illustration of the Prism block is shown in Fig. \ref{fig:prism}, the input signal is first projected and separated and then modulated by $\pi$-RoPE to capture different features at different scales, and finally processed via differential de-noising to remove the noise. The incoherence between $U_s$ and $U_n$ ensures that the denoising process suppresses high-frequency interference without eroding the integrity of the semantic signal accumulated in the residual stream.

For multi-head attention, we just introduce subspaces $[\mathbf{U}_{s,1},\dots,\mathbf{U}_{s,K}]$ and $[\mathbf{U}_{n,1},\dots,\mathbf{U}_{n,K}]$ for clean signal and noise respectively. Under the same procedure, we can obtain the multi-head attention directly:

\begin{equation}
  \begin{split}
\mathbf{Z}^{\ell+1} &\approx \mathbf{Z}^{\ell} + \eta \sum_k \left\{  \mathop{\text{Att}}(\mathbf{U}_{s,k},\mathbf{Z}) (\mathbf{Z}^\ell \mathbf{U}_{s,k}) \mathbf{U}^\top_{s,k} \right. \\
    & ~~~\left.  - \lambda \mathop{\text{Att}}(\mathbf{U}_{n,k},\mathbf{Z}) (\mathbf{Z}^\ell \mathbf{U}_{n,k}) \mathbf{U}_{n,k}^\top \right\},
  \end{split}
  \label{eq:gradient_ascent}
\end{equation}
where $\mathop{\text{Att}}(\cdot,\cdot)$ is the attention matrix. 

While in our derivation, MCR$^2$ formulates representation learning as separating signal from noise, this may not be applied directly for natural language processing (NLP) tasks. We instead reinterpret the signal as semantic subspace corresponding to long-range causal dependency and the noise as syntactic subspace with respect to local constraints, rigid structural artifacts, repetition requiring subtraction. The noise removing process is now reinterpreted as suppressing interference to prevent high-frequency syntactic artifacts (e.g., repetitive loops) from spectrally overshadowing the low-frequency semantic signals. Note that this subtraction operates on the residual update $\Delta Z$, filtering the attention head's contribution, rather than erasing the syntactic information already accumulated in the residual stream. Under this definition, the 'Attention Sink' phenomenon is classified as a high-energy syntactic artifact residing in the noise subspace.

In practice, the underlying subspaces $\mathbf{U}_s,\mathbf{U}_n$ are unknown and must be learned end-to-end via backpropagation. To eliminate the variance explosion, and considering the fact that the dictionary can be overcomplete, we suggest choosing $\eta = \frac{1}{\sqrt{R K}}$ where $R$ is the expansion ratio of the dictionary and input space \footnote{In standard Transformer the ratio is $1$, and we suggest setting $R=2$ to ensure same number of parameters between Prism and standard Transformer.}.  From this perspective, we can view weight optimization as finding the subspaces $\mathbf{U}_s$ and $\mathbf{U}_n$ maximizing the coding rate reduction and separate noise from noisy signal. 

\subsection{Mean-Field Dynamics and Spectral Decoupling}
Considering a one-layer network with Prism block, as the number of heads $H \to \infty$, the summation becomes integral and the output of the Prism layer reduces to 

\begin{equation*}
  f(z) = z + \int \Phi(u,z) \rho_S(u) \, \mathrm{d}u -\lambda \int \Phi(u,z)\rho_N(u) \, \mathrm{d}u
\end{equation*}
where $\rho_S(u)$ and $\rho_N(u)$ are the probability density function of the weights $u$, and the kernel function $\Phi(x,z)$ is the kernel with respect to the attention matrix. Although practical models use a finite number of heads, the mean-field formulation provides insights into the asymptotic behavior of the signal/noise separation dynamics.

Considering the mean square error with training data $y,z$,
$$ \mathcal{L}(\rho_S, \rho_N) \approx \frac{1}{2} \mathbb{E}_{z,y}| y - (\text{Signal} - \lambda \cdot \text{Noise}) |^2, $$ 
under the mean-field limit, the Wasserstein gradient flow is given by:
\begin{align*}
    \partial_t \rho_S &= \nabla_u \cdot \left( \rho_S \nabla_u \left( V_{data}^S(u) + \int K_{SS} \, \mathrm{d} \rho_S \right.\right. \\ 
     &~~~\left.\left.-\lambda \int K_{SN} \, \mathrm{d}\rho_N \right) \right) \\
    \partial_t \rho_N &= \nabla_u \cdot \left( \rho_N \nabla_u \left( V_{data}^N(u) - \lambda \int K_{NS} \, \mathrm{d}\rho_S  \right.\right. \ \\ &~~~+\left.\left. \lambda^2 \int K_{NN} \, \mathrm{d}\rho_N \right) \right)
\end{align*}

where $K(u,v) = \int \Phi(u,z)\Phi(v,z) \rho_z \, \mathrm{d}z$ is the kernel function characterizing the coherence or similarity between different heads and the $V_\mathrm{data}^S(u)$ and $V_\mathrm{data}^N(u)$ are terms with respect to training data, and $K_{SN},K_{NN},K_{NS}$ are the cross-interaction terms between signal and noise heads.

In sequential modeling, the time step $t$ is usually embedded by RoPE. Under RoPE, each head has its own frequency. Using the properties of RoPE $\langle R_t q,R_{t-\tau}k\rangle = q^\top R_\tau k$ where $R_t$ is the rotation matrix defined by RoPE and $\tau$ is the time step interval. Motivated by this, we thus decouple the pre-softmax logit for head $u$ by semantic part and position part:
\begin{equation}
  S_u(\tau) = C_u \cdot \cos (\omega_u \tau + \phi_u),
\end{equation}
where $C_u$ is some function depending on $u$ representing content. Due to the fact that the position and absolute content are usually decoupled, and each output is auto-regressive thus we approximate the kernel function by:
\begin{equation}
  K(u,v) \sim \mathbb{E}[C_u C_v] \cdot \mathbb{E}_\tau [\cos \omega_u \tau \cdot \cos \omega_v \tau ],
\end{equation}
where $\mathbb{E}_\tau$ is the integral over different steps $\tau$. 

The second term can be written as:
\begin{equation*}
  \cos \omega_u \tau \cdot \cos \omega_v \tau = \frac{1}{2} \cos((\omega_u + \omega_v)\tau) + \cos((\omega_u - \omega_v)\tau) 
\end{equation*}

The first term is oscillatory and vanishes as $\tau \xrightarrow \infty$ due to phase cancellation (destructive interference) over long contexts, analogous to the Riemann-Lebesgue lemma. Resonance occurs when the frequencies between signal and noise heads align, leading to coupling. To eliminate this, we want the orbit to be dense and uniformly distributed on the torus.  Motivated by Weyl's Equidistribution Theorem \cite{einsiedler2011ergodic}, we suggest choosing the irrational frequency in RoPE. However, mere irrationality may be insufficient; we require not only non-resonance but also that the system is far from resonance. This requirement is analogous to the stability conditions in Hamiltonian mechanics, specifically the Kolmogorov-Arnold-Moser (KAM) theorem \cite{arnold1989mathematical}. 

To see this, we note that the cross-interaction kernel $K_{SN}$ and $K_{NS}$ explicitly quantifies the violation of subspace incoherence. While MCR$^2$ requires incoherence ($\mathbf{U}_S^\top \mathbf{U}_N \approx 0$), RoPE can induce subspace alignments dynamically. In the perturbative series analysis aimed at eliminating these interaction terms  $K_{SN}$ (via canonical transformation), denominators of the form $\frac{1}{n\omega_S - m\omega_N}$ appear (where $n,m$ are integers). Standard RoPE frequencies lead to the so-called \textit{Small Divisors Problem} (resonance), causing the perturbative series to diverge. Physically, this implies that the interaction terms ($K_{SN},K_{NS}$) become \textit{non-removable}, effectively mixing the signal and noise subspaces over long contexts and violating the MCR$^2$ objective. 

To avoid resonance and divergence, we finally suggest using irrational numbers satisfying Diophantine conditions \cite{einsiedler2011ergodic} in RoPE. This leads to our $\pi$-RoPE \footnote{One may choose $\phi = \frac{1+\sqrt{5}}{2}$, but the number may be too small to separate the long-range and short-range signal.} 
\begin{equation}
  \theta_S =  \pi \theta_{\mathrm{base}}  \quad \theta_N = \frac{1}{\pi} \theta_{\mathrm{base}},
\end{equation}
where $\theta_{\mathrm{base}}$ is the base frequency \footnote{Usually denoted as $b$ with typical value $10000$.}  in RoPE that is usually chosen to be $10000$ in GPT-2. More theoretical benefit can be shown, if we study the Wasserstein gradient flow of Prism with respect to $\rho_S$ and $\rho_N$, their evolution equation has an interaction kernel term which can be eliminated by $\pi$-RoPE. Thus, the subspaces between signal and noise are nearly decoupled and the whole optimization becomes a good optimization problem. The signal heads find the long-range and low-frequency signal and the noise heads fit the high-frequency, short-range solution. 

Finally, we adopt $\pi$-RoPE as an ansatz to design our rotary embeddings and give a conjecture below:

\textbf{Conjecture:} (Non-Resonance Condition in Prism). We posit that selecting a rotational frequency ratio $\omega_S/\omega_N \notin \mathbb{Q}$ (an irrational number, specifically irrational numbers satisfying Diophantine conditions such as $\pi$ or $\phi$) minimizes the integrated cross-correlation $K_{\text{SN}}, K_{\text{NS}}$ between the signal and noise subspaces over long sequences.

\section{Experiments}
\subsection{TinyStories Dataset}
To validate the theoretical hypothesis of signal-noise decomposition, we conducted experiments on the TinyStories dataset \cite{eldan2023tinystories}, which is a benchmark designed to test reasoning and coherence in small-scale language models. We deliberately choose a compact setup to conduct controlled intervention analysis, isolating the geometric mechanism from the interference of massive parameter redundancy.

Model Configuration: We implemented a "Prism-mini" variant and a GPT-2 baseline. We train a 8-layer Prism with overcomplete attention ($R=2$, resulting in $16$ physical heads per layer). The heads are split into pairs of signal head and noise head

RoPE Configuration: We applied $\pi$-RoPE for frequency separation. Signal heads utilized a base frequency $\theta_S=10000\pi \approx 31415$, while Noise heads utilized $\theta_N=10000/\pi \approx 3183$.

Training Details: We implement the Prism via nanoGPT \footnote{\url{https://github.com/karpathy/nanoGPT/tree/master}} and then train the Prism-mini from scratch on a single NVIDIA RTX 4090 GPU for $20000$ steps with a batch size of $32$ and context length of $512$. We utilized the AdamW optimizer with cosine learning rate decay. The learning rate is 6e-4, and the warmup step is $1000$. We use the cosine schedule from $0.01$ to $0.1$ for hyperparameter $\lambda$ to stabilize training.

\begin{figure}[t]
  \centering
  \includegraphics[width=\linewidth]{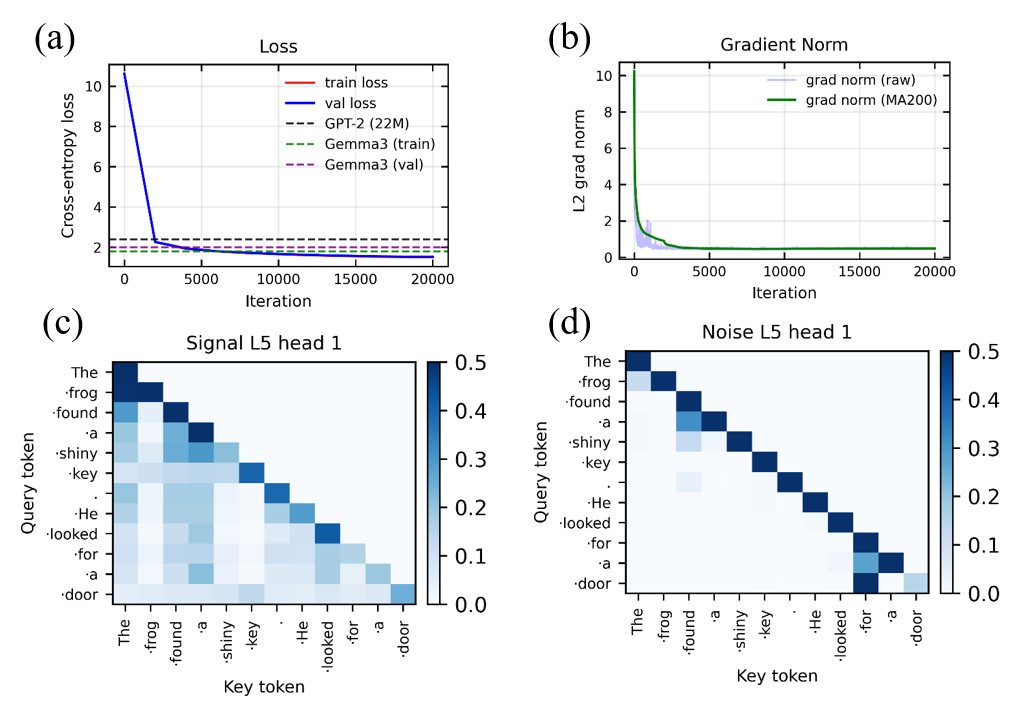}
  \caption{\textbf{Training Dynamics: Emergent Functional Specialization} 
    \textbf{(a)} Training and validation loss curves on the TinyStories dataset. 
    \textbf{(b)} Evolution of the L2 Gradient Norm. 
    \textbf{(c)-(d)} Comparative visualization of attention maps in Layer 6, Head 2. 
    \textbf{(c) Signal Head:} Driven by low-frequency $\pi$-RoPE, it captures sparse, long-range semantic dependencies (e.g., attending to specific entities). 
    \textbf{(d) Noise Head:} Driven by high-frequency $\pi$-RoPE, it captures local syntactic artifacts and background noise. }
  \label{fig:training}
\end{figure}

The training dynamics and the visualization of attention matrix is shown in Fig. \ref{fig:training}. The Fig. \ref{fig:training} (a) indicates that Prism (50M) exhibits rapid convergence, achieving a validation loss of $\approx 1.55$ and surpassing the GPT-2 (22M) baseline (red dashed line \footnote{\url{https://huggingface.co/jonasknobloch/gpt2_u020_tiny-stories_1024_dpos}}), demonstrating high parameter efficiency even better than Gemma 3 (dashed lines in blue, purple, and green \footnote{\url{https://github.com/di37/gemma3-270M-tinystories-pytorch}}).  Despite the potential instability of the differential operator ($S - \lambda N$), Fig. \ref{fig:training} (b) shows that the training is stable, and the gradient norm stabilizes at a low magnitude ($\approx 0.5$), indicating a smooth optimization landscape facilitated by the white-box constraints. Specific Visualization in Fig. \ref{fig:training} (c) and Fig. \ref{fig:training} (d) indicates that the signal heads of Prism automatically focus on the long-range semantic link (such as the causal link between "door" and "key"), whereas the noise heads focus more on local syntax (e.g. dependency between "for" and "door").

\begin{figure}[h]
    \centering
    \includegraphics[width=\linewidth]{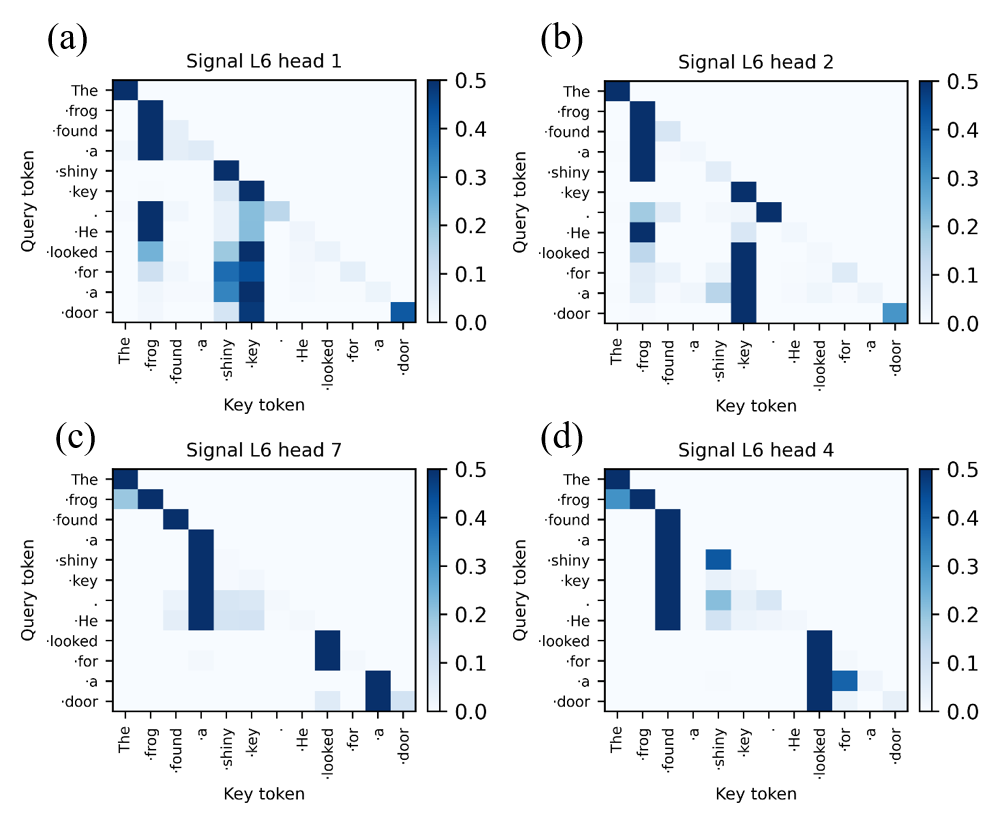}
    \caption{\textbf{Overcomplete Signal Subspace Visualization:} 
    \textbf{(a) Subject-centric:} Maintains a persistent vertical attention band on the word ``\textit{frog}'', tracking the narrative protagonist.
    \textbf{(b) Object-centric (Causal):} When predicting the target token ``\textit{door}'' (last row), this head explicitly attends to the distal causal antecedent ``\textit{key}'', bridging the logical gap between the tool and the target.
    \textbf{(c) Structural:} Attends to the determiner ``\textit{a}'', likely tracking the introduction of new entities.
    \textbf{(d) Action-centric:} Anchors the sequence to the primary predicate ``\textit{found}'' and ``\textit{looked}'', tracing the action.}
    \label{fig:intra_layer_specialization}
\end{figure}

To visualize the learned signal, we visualize the attention patterns of four distinct heads of the sixth layer from the signal head group ($\pi$-scaled RoPE) for the input sequence ``\textit{The frog found a shiny key... looked for a door}''. As shown in Fig.~\ref{fig:intra_layer_specialization}, despite sharing the same low-frequency inductive bias, the overcomplete basis ($R=2$) enables the emergence of diverse semantic roles without supervision and reveals the richness within the signal subspaces. Thus, the overcomplete parameterization allows the model to allocate dedicated heads for specific semantic roles such as subject and action, effectively revealing a sparse dictionary learning process.

\subsection{OpenWebText Dataset}
\begin{figure}
    \centering
    \includegraphics[width=0.95\linewidth]{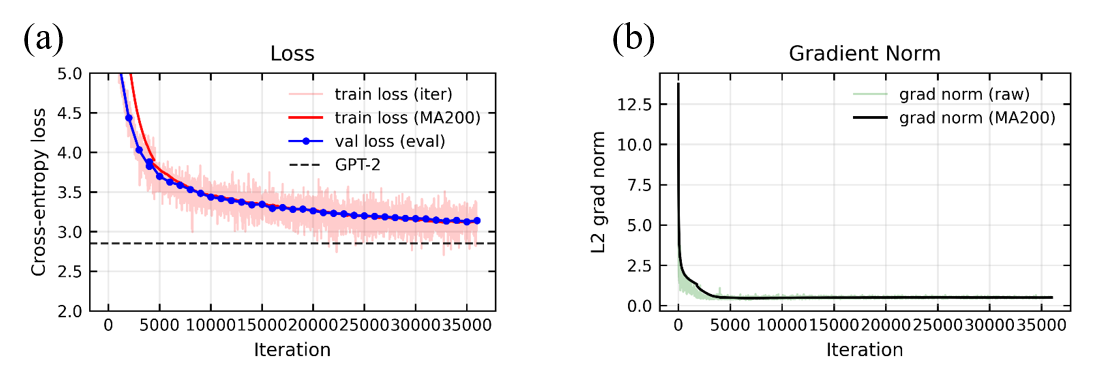} 
    \caption{\textbf{Scale-up Validation of Prism (124M parameters) on OpenWebText.} \textbf{(a) Convergence:} Training (red) and validation (blue) loss curves over 36k iterations (approx. 18B tokens). Prism demonstrates a healthy, monotonic convergence trajectory, reaching a final validation and training loss around $3.10$, marginally higher than the GPT-2 baseline (dashed line, $\approx 2.85$ ) without extensive hyperparameter tuning. \textbf{(b) Dynamical Stability:} The $L^2$ Gradient Norm (green) and its moving average (black) exhibit an exceptionally stable behavior, rapidly settling to a constant baseline ($\approx 0.5$) after the initial learning-rate warmup stage. The absence of gradient spikes or collapse confirms the numerical robustness of the proposed differential denoising operator $S - \lambda N$ in high-dimensional optimization landscapes.}
    \label{fig:Dynamics OpenWebText 124M}
\end{figure}
\paragraph{Optimization Landscape and Gradient Stability}
The transition from the simple TinyStories data to the high-entropy, long-tail OpenWebText \cite{Gokaslan2019OpenWeb} data serves as a stress test for the training validity of the Prism architecture. Standard Transformers, when scaling up or encountering complex data distributions, often suffer from training instability (e.g., loss spikes or gradient spikes).

In contrast, Fig. \ref{fig:Dynamics OpenWebText 124M}(b) reveals an intrinsic dynamical stability property of Prism. Following the initial warmup phase (\~2k steps), the $L^2$ gradient norm may act as a "thermodynamic invariant," stabilizing at $\sim 0.5$ with reasonable variance. This behavior is non-trivial. It may suggest that the white-box constraints—specifically the de-correlation via $\pi$-RoPE and the subtractive denoising operator $S-\lambda N$ effectively smooths the optimization landscape of Prism. Thus, Prism does not struggle to balance the signal and noise subspaces. In addition, the geometry of the architecture may naturally guide the gradient flow along a smooth manifold, preventing the "rugged" loss surfaces typical of unconstrained black-box models.

\paragraph{Comparative Convergence Performance}
As shown in Fig. \ref{fig:Dynamics OpenWebText 124M}(a), Prism achieves a final validation loss around 3.10 after 2 epochs on the OpenWebText dataset (around 18B tokens). Although there remains a marginal gap compared to the heavily optimized GPT-2 baseline ($\approx$ 2.85), this result is highly encouraging for two reasons:
\begin{enumerate}
    \item Zero-Tuning: The reported performance was achieved using a "first-shot" configuration, inheriting hyperparameters directly from the TinyStories setup without a careful sweep. Standard Transformers are notoriously sensitive to learning rate and initialization; Prism's ability to converge competitively "out-of-the-box" indicates high transferability.
    \item Effective Capacity: The loss curve in Fig. \ref{fig:Dynamics OpenWebText 124M}(a) shows little sign of saturation or overfitting (the gap between training and validation loss remains healthy). This implies that the white-box architecture does not overly constrain the model's expressivity. The Prism neural network is still successfully compressing the semantic content of the information in OpenWebText, validating that the "signal" subspace is sufficiently high-dimensional to capture open-domain knowledge.
\end{enumerate}

The successful training of Prism-124M on OpenWebText confirms that the physical and geometric interpretability mechanisms do not come at the cost of training stability or representational collapse at scale.

\subsection{Emergence of Induction Primitives in Prism-124M and Structure-Content Decoupling}
While the 124M parameter regime is insufficient for storing extensive world knowledge, our analysis of Prism-124M consistently exhibits ``Structure-Content Decoupling". Unlike standard Transformers trained with cross-entropy, which often degrade into incoherence under uncertainty, Prism exhibits a strong inductive bias towards preserving the form of data even when the content is lost. We interpret this as a direct consequence of the MCR$^2$ objective, which prioritizes the compression of low-rank, high-occurrence structural patterns (Syntax) into orthogonal subspaces before the emergence of high-rank and complex semantic manifolds.

We categorize this emergence into three ascending tiers of induction primitives with three cases:
\begin{tcolorbox}[width=\linewidth]
\textbf{Prompt:} CLICK HERE to download. [AD] Buy now! The capital of France is Paris. [AD] Free shipping. The capital of Germany is  \\
\textbf{Output:} Germany.
\label{Case: Germany}
\end{tcolorbox}
As shown in the case above, when prompted with "The capital of Germany is", the Prism model, lacking the specific knowledge of "Berlin", does not hallucinate a random entity (e.g., "Paris") but instead outputs "Germany".
This represents the most primitive form of In-Context Learning (ICL): the naive copy head. From an information-theoretic perspective, the MCR$^2$ objective discourages the introduction of noise. Consequently, the attention mechanism retreats to the signal with the highest local saliency—the token "Germany" itself. This low-entropy replication may indicate that Prism has successfully established basic retrieval circuits, prioritizing signal preservation over fabrication.

\begin{tcolorbox}[width=\linewidth]
\textbf{Prompt:} \verb|{"user": {"id": 123, "attri|\\ \verb | butes":  ["admin", "editor" | \\
\textbf{Output:}\verb|]}}, "text_type":|
\label{Case: Structural Induction}
\end{tcolorbox}
We test the model by the "bracket closure" test, when generating a JSON object, the model correctly identifies the truncation and immediately generates \verb|]}}| to close the structure (syntactic recovery). This shows the syntactic induction. This is the defining example of Structure-Content Decoupling. The model has learned the topology and the main structure of the JSON language manifold (open brackets must be closed), but lacks the capacity to fill its volume (diverse content), indicating that the syntax subspace is robust and orthogonal, while the semantic subspace remains under-populated.

\begin{tcolorbox}[width=\linewidth]
\textbf{Prompt:}The scientist discovered a new species of blue butterfly in the deep Amazon rainforest. Upon closer inspection, they realized that  \\
\textbf{Output:} the caterpillars had a distinctive characteristic, indicating that the caterpillar was in fact a species of butterfly that was not a butterfly. However, the caterpillars also ... The research was published in the scientific journal Monthly Notices of the Royal Society.

The study was presented this week at the annual meeting of the American Association for the Advancement of Science, held in Hawaii.
\label{Case: Manifiold Projection}
\end{tcolorbox}
When generating scientific text, the model has logical collapse but strictly adheres to the tone, citation format (``Monthly Notices of the Royal Society"), and complex sentence structures of academic writing. This confirms our Structure-Content Decoupling hypothesis: The MCR$^2$ objective successfully compresses the rigid forms of scientific discourse into the model's signal subspaces, while the dense facts of biology require a larger model capacity to resolve. This evidences a clear topological acquisition preceding semantic filling. Prism knows how to write a science paper, it just doesn't know science yet.

\section{Analysis of Repetitive Attractors}
To analyze the repetitive behavior of Prism and compare it with the standard Transformer network (GPT-2), we design an experiment and give an analysis of $\pi$ rotary position embedding (RoPE) based on Hamiltonian dynamics.

Following the paradigm of mechanistic interpretability, we utilize compact models (124M) as controlled testbed to isolate geometric dynamics from interference of massive parameter redundancy.

\subsection{Experiment Setup}
Standard autoregressive Transformers frequently exhibit degenerative looping pathologies, a phenomenon theoretically linked to the self-reinforcing activation of induction heads and the Attention Sink \cite{Xiao2024}. In such systems, the recurrence of a pattern typically deepens the local potential well of the attractor, rendering the loop increasingly stable and difficult to escape. We propose that Prism, governed by white-box geometric constraints, introduces a countervailing ``restorative force.'' This experiment aims to empirically isolate this mechanism, distinguishing between the \textit{scale-invariant collapse} of standard attention and the hypothesized \textit{repetition fatigue} inherent to our architecture.

We formulate the evaluation as a stability analysis of the generation manifold under varying thermal noise. We employ a high-frequency cyclical trigger, defined as the prompt $P_N = (s)^N$, where the atomic unit $s = \texttt{A little alarm clock, }$ and $N \in \{2, 5, 10, 20, 50\}$ represents the repetition count.
For both the Prism-124M and the GPT-2 baseline, we determine the \textit{Critical Temperature} $T_{c}(N)$, defined as the minimum sampling temperature required for the model to successfully diverge from the repetitive loop within the next 20 generated tokens (escape rate $\ge 50\%$).
Here, $T_{c}$ serves as a proxy for the energy barrier height of the attractor. We choose the sampling temperature to vary from 0.1 to 1.2 and detect the loop by matching the semantic word ``alarm clock".

\begin{figure}[ht]
    \centering
    \includegraphics[width=\linewidth]{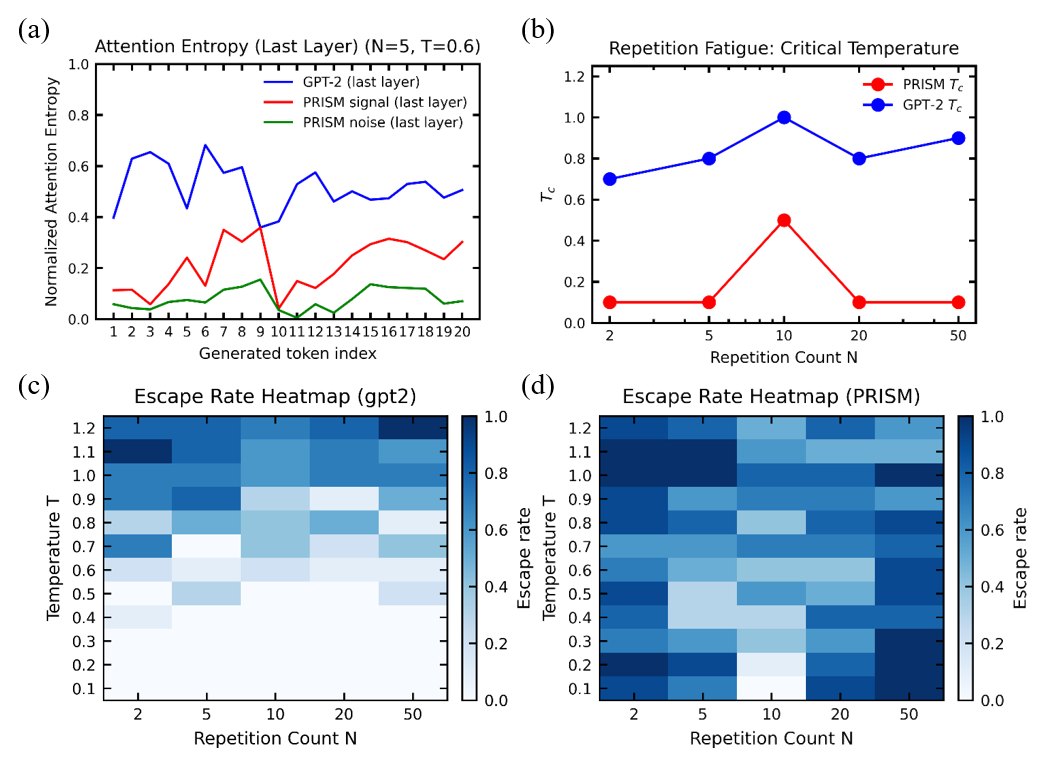}
    \caption{\textbf{Thermodynamic Analysis of Generation Dynamics: Evidence of Restricted Ergodicity.} 
(a) \textbf{Entropy Evolution (Last Layer):} During autoregressive generation, standard GPT-2 (blue line) exhibits anomalously high entropy ($\approx 0.6$), indicating that the model fails to anchor onto specific semantic tokens and suffers from uncertainty and repetition. In contrast, Prism's signal head (red line) and noise head (green line) maintains low entropy ($\approx 0.2, 0.1$ respectively), demonstrating \textbf{structural sparsity} and precise semantic locking at the output stage.
(b) \textbf{Critical Temperature of Repetition:} Critical temperature $T_{c}$ as the minimal sampling temperature required to escape a local repetition loop. GPT-2 requires high thermal noise ($T \approx 0.8\text{-}1.0$) to break resonance, indicating deep metastable traps (Arnold Tongues). In contrast, Prism spontaneously escapes even at near-zero temperatures ($T \approx 0.1$), suggesting intrinsic restricted ergodicity.
(c)-(d) \textbf{Escape Rate Heatmaps:} Prism (d) retains high escape probabilities (dark blue) across the low-temperature regime compared to GPT-2 (c), validating that Prism's generation diversity stems from geometric stability rather than thermodynamic fluctuation induced by temperature.}
\label{fig:ergodicity}
\end{figure}

\subsection{Hamiltonian Analysis of Resonance and Attention}
To analyze the stability of long-context attention, we introduce a KAM-inspired dynamical analysis motivated by an isomorphism to Hamiltonian dynamics, where we view the transformer's forward pass as a classical Hamiltonian system \footnote{While the Transformer is formally a dissipative system due to softmax, the resonance dynamics of the pre-softmax logits can be effectively modeled as a local Hamiltonian system. See Discussion for details.}. We identify the token position index as discrete time $t$, and the Rotary Positional Embedding (RoPE) as a rotation operator on a torus $\mathbb{T}^{D} (D=d/2)$, where $d$ is the input dimension of tokens.

We define the canonical coordinates ($\phi, I$) for the embedding space, where $\phi$ represents the phase angles (for position $m$, the position is $\phi_j(m) = m\theta_j$ where $\theta_j$ is the frequency required by RoPE and $j=1,...,d/2$, indexing the pairs.) and $I$ represents the action variables (encoding token amplitude, $I_j = \frac{1}{2}(q_{2j}^2 + q_{2j+1}^2)$). For convenience, we denote $\mathbf{\omega} = (\theta_1,...,\theta_D)$ as the collection of frequencies. The effective Hamiltonian $H$ of the attention mechanism can be decomposed into an integrable free term $H_0$ and a perturbation potential $V$:
\begin{equation}
    H(\phi,I) = H_0 + \epsilon V,
\end{equation}
where $H_0 = \sum_{j=1}^{d/2} \omega_j I_j $ is the integrable part for ideal rotation. The interaction term $V$ is the pre-softmax logit $\mathrm{Re}(\mathbf{q} \cdot \mathbf{k})$ coupling all the angles. We can write the logit as Fourier series: $$V = \sum_{\mathbf{k} \in \mathbb{Z}^{D} \backslash \{0\}} V_\mathbf{k}(I) e^{i \mathbf{k} \cdot {\bm\omega}},$$
where $\mathbf{k}=(k_1,...,k_D)$ represents the integral indicating the coupling of frequency with respect to different dimensions. Here, the scaling factor $1/\sqrt{d_k}$ serves as the small physical coupling constant $\epsilon$ keeping the system in the near-integrable regime and the perturbation analysis reasonable.

The standard way to analyze the stability of the Hamiltonian is canonical transform, aiming at cancel the perturbation term $\epsilon V$ such that the Hamiltonian only depends on the new variable $(\psi,J)$ defined as 
\begin{equation}
    \begin{split}
        I = J + \epsilon \frac{\partial S_1}{\partial \phi} \\
        \psi = \phi + \epsilon \frac{\partial S_1}{\partial J}, \\
    \end{split}
\end{equation}
where $S_1$ is the so-called generation function.

After the transformation, the Hamiltonian should depend only on the new variable $J$:
\begin{equation}
    H(\phi,I) = H_0(J + \epsilon \frac{\partial S_1}{\partial \phi}) + \epsilon V(\phi,J) = K(J).
\end{equation}

\subsection{Resonance and Arnold Tongues: The Origin of Attention Sinks}
The stability of this system under perturbation is governed by the Small Divisor Problem. The generating function eliminating the interaction term involves Fourier coefficients of the form:
\begin{equation}
    S_\mathbf{k} \propto \frac{i V_\mathbf{k}}{\mathbf{k}\cdot {\bm \omega}},
\end{equation}
where $\mathbf{k}\in \mathbb{Z}^D$ is the resonance vector. Resonance occurs when the denominator $\mathbf{k}\cdot {\bm \omega} \approx 0$, causing the perturbation series to diverge. In the phase space of circle maps, these resonance regions manifest as Arnold Tongues.

\textbf{The Pathology of Geometric Progressions:} Standard Transformers utilize a geometric progression for frequencies $(\theta_j=b^{-\frac{2j}{d}})$. We postulate that these frequencies possess strong arithmetic properties, creating dense networks of low-order resonances.

\textbf{Repetition Loops:} Fig. \ref{fig:ergodicity} provides a thermodynamic diagnosis of the 'repetition fatigue' pathology. As illustrated in Fig. \ref{fig:ergodicity}(b), the standard Transformer (GPT-2) exhibits a characteristic of mode-locking: it requires significant external thermal noise ($T_c \approx 1.0$) to kick the system out of repetition loops. In dynamical systems terms, the model's trajectory is trapped in resonance islands (or Arnold Tongues) induced by the rational structure of standard RoPE frequencies. Conversely, Prism demonstrates intrinsic restricted ergodicity on the maximal tori (quasi-periodicity). By enforcing spectral irrationality via $\pi$-RoPE, Prism destroys these resonance islands. The model maintains a quasi-periodic trajectory that naturally explores the phase space, evidenced by its ability to escape loops even in the deterministic limit $(T\rightarrow 0)$. This shows that Prism's superior generation quality stems from geometric stability, not merely better probability calibration.

\begin{figure}[H]
    \centering
    \includegraphics[width=\linewidth]{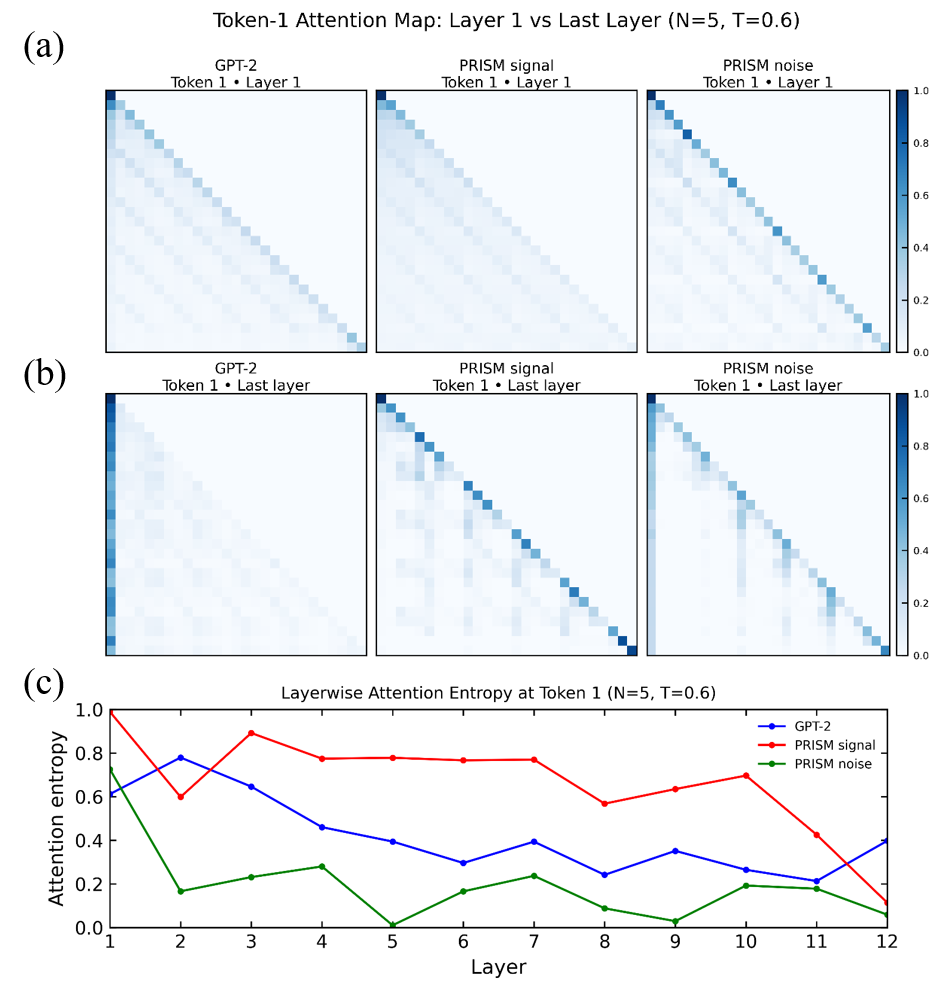} 
    \caption{\textbf{Mechanism of Attention Stability: Elimination of Rank Collapse via Signal/Noise Separation.} 
\textbf{(a-b): Evolution of Attention Maps.} 
(a) \textbf{Standard GPT-2:} At Layer 1 (top-left), attention is distributed reasonably. However, by the Last Layer (bottom-left), the model suffers from the pathological ``Attention Sink'' phenomenon, dumping massive probability mass onto the initial token (dark blue vertical band) to satisfy softmax normalization, indicating a collapse of semantic resolution.
(b) \textbf{Prism Architecture:} The signal stream (center column) maintains clean, diagonal semantic dependencies from Layer 1 to the Last Layer, free from sinks and reveals clear periodic attention structure. The noise stream (right column) detects the ``Attention Sink" by viewing it as a high-frequency syntactic noise (vertical bands) into a noise subspace orthogonal to signal subspace.
\textbf{(c): Layerwise Entropy Dynamics.} This plot quantifies the \textit{Rank Collapse}. GPT-2 (blue line) shows a nearly monotonic decay in (normalized) attention entropy, indicating a progressive loss of channel capacity (feature rank) with depth. Prism's signal stream (red line) achieves \textit{Isentropic Flow}, maintaining high information bandwidth even in the deep layers ($L=10$). }
\label{fig:mechanism}
\end{figure}

\textbf{Attention Sinks:} We view the 'Attention Sink' phenomenon not as a necessary feature for streaming stability, but as a symptom of feature rank collapse. As shown in Fig. \ref{fig:mechanism}(c), the attention entropy of standard GPT-2 (blue) decays with depth. By the final layers, the model has effectively degenerated into a low-rank operator, forced to attend to the trivial 'sink token' because it has lost the capacity to resolve complex semantic signals.

\subsection{Resonance and Arnold Tongues: The Origin of Attention Sinks}
To restore stability, we draw inspiration from the Kolmogorov-Arnold-Moser (KAM) Theorem \cite{arnold1989mathematical}. The theorem states that invariant tori (stable quasi-periodic orbits) persist under perturbation if the frequency vector ${\bm \omega}$ satisfies the Diophantine Condition: $\mathbf{\mathbf{k}\cdot {\bm \omega}} \geq \gamma \|\mathbf{k}\|^{-\tau}$, where $\gamma,\tau>0$ are real numbers.

As shown in Fig. \ref{fig:mechanism}, Prism fundamentally alters this trajectory through spectral separation. The architecture bifurcates the information flow: coherent semantics are preserved in the signal stream (red), which exhibits \textit{Isentropic Flow} (near constant entropy across layers), while incoherent artifacts are shunted to the noise stream (green). The noise stream effectively serves as a detector and shield, protecting the signal stream from rank degeneration and falling into Attention Sink. This geometric decoupling allows Prism to operate without ad-hoc sink tokens, offering a first-principles solution to the instability of deep transformers.

Analogous to the predictions by KAM theory, the presence of irrational frequencies encourages quasi-periodicity, where trajectories densely cover the invariant tori without collapsing into periodic orbits. As shown in Fig. \ref{fig:ergodicity}(b), Prism exhibits a lower critical temperature $T_c$ compared to GPT-2, suggesting a non-resonant dynamics. This implies that the system effectively avoids mode-locking (Arnold Tongues), maintaining rich representational diversity rather than being trapped in low-complexity limit cycles.

We acknowledge that scaling the base frequency by $\pi$ (resulting in $b\approx 3.14 \times 10^4$) also increases the characteristic wavelengths, which is known to aid long-context extrapolation. However, our Hamiltonian analysis suggests that the arithmetic property (irrationality) provides a distinct protection in preventing resonance beyond mere scaling. Disentangling the contribution of magnitude scaling from spectral irrationality remains a key objective for future work.

\subsection{Implications for Large Scale Models: KAM-RoPE for Standard Transformers} 
We hypothesize that the ``Attention Sink" phenomena in current Large Language Models fundamentally stem from the spectral entanglement between global semantic signals and local syntactic noise. Our theoretical analysis suggests that the ``Attention Sink" or repetitive loop phenomenon in standard LLMs may be fundamentally viewed as a physical resonance phenomenon arising from the strict geometric progression of RoPE frequencies, $\theta_j=b^{-2j/d}$. This progression possesses strong arithmetic coherence, creating dense networks of Arnold Tongues (resonance traps).

While Prism solves this structurally via $\pi$-RoPE, we propose a lightweight plug-and-play intervention for existing pre-trained models, termed \textbf{KAM-RoPE}. The core idea is to disrupt the strict geometric progression of the frequency lattice via spectral detuning:
$$\theta_j = \theta_j (1+\xi_j) \quad \xi_j \sim U(-\epsilon,\epsilon).$$

Here, $\xi_j$ is a quenched disorder term (sampled once and fixed) that acts as a ``spectral jitter." Even a minimal perturbation $(\epsilon \approx 1e-4)$ is theoretically sufficient to violate the low-order resonance conditions required for mode-locking and rigorous geometric series, pushing the system towards KAM stability without destroying the global relative position semantics.

Additionally, we recommend modifying the frequency base b with an irrational scaling factor (e.g., the Golden Ratio $\phi= \frac{1+\sqrt{5}}{2}$ ):
    $$b'=b\cdot \phi \approx 10000 \times 1.618\dots.$$
    
The Golden Ratio is chosen because it is the most irrational number (in the sense of Diophantine approximation), theoretically providing the strongest protection against Arnold Tongues. We hypothesize that fine-tuning standard Transformers with this KAM-RoPE configuration may significantly expand their effective context window and suppress generation loops, serving as a direct validation of the resonance theory.

Finally, we distinguish the representational entropy (internal layers) and the decision entropy (final layer). High representational entropy (Fig. \ref{fig:mechanism}) prevents information loss (rank collapse), while low decision entropy (Fig. \ref{fig:ergodicity}(a)) indicates confidence and sparsity in the final token prediction. Prism succeeds in both regimes, whereas GPT-2 fails in both (collapsing internally, yet remaining uncertain at the output).

\section{Conclusion, Discussion and Future Works}
In this work, we derived the Prism layer from the first principle of maximizing coding rate reduction, demonstrating that semantic reasoning and syntactic processing are spectrally separable regimes within the Transformer architecture. By imposing an overcomplete dictionary and enforcing spectral separation via $\pi$-RoPE, Prism achieves unsupervised functional disentanglement, evidenced by the emergence of distinct, specialized functional heads and the effective isolation of Attention Sinks by introducing the additional noise head. In other words, our architecture effectively compartmentalizes the Attention Sink into the syntactic (noise) stream (as seen in Fig. \ref{fig:mechanism}(b)), verifying that the sink is a necessary artifact of softmax that should be isolated rather than eliminated.

Limitations of this study include the restricted parameter scale (50M, 124M). However, the Prism architecture retains high compatibility with modern computational primitives such as FlashAttention \cite{Dao2022FlashAttentionFA} and sparse scaling mechanisms like Mixture of Experts (MoE) \cite{Shazeer2017OutrageouslyLN}. The gradient norm of prism is stable across toy datasets like Tinystories and real datasets like OpenWebText, revealing signals of scalability. We are currently extending the Prism framework to larger-scale datasets and deeper parameterizations to verify the persistence of these geometric properties at scale and the scalability of Prism. Scaling experiments on large models are currently underway on clusters.

Theoretically, future work will focus on establishing a rigorous derivation of the mean-field dynamics under the overcomplete setting and focus on establishing bounds for the non-resonance condition for the transcendental $\pi$-RoPE basis. Furthermore, we aim to extend the white-box paradigm to the Feed-Forward Networks (FFN) via sparse coding unrolling. We believe that the FFN layer can be understood and replaced by a finite-step approximation of sparse coding optimization, effectively refining the signal by projecting it onto a union of low-dimensional subspaces.

\paragraph{Limitations of Hamiltonian Analysis:} We acknowledge that the Transformer forward pass, strictly speaking, is a dissipative dynamical system due to the contractive nature of LayerNorm and Softmax operators. In such systems, phase space volume is not conserved, and trajectories typically converge to attractors (fixed points or limit cycles) rather than preserving invariant tori. 

However, our Hamiltonian analysis focuses specifically on the pre-softmax logits dynamics. The interaction term $\mathrm{Re}(\mathbf{q}\cdot \mathbf{k})$ describes a phase coupling mechanism that is fundamentally oscillatory. While dissipation governs the amplitude of the signal, the phase synchronization (which leads to repetition loops and attention sinks) is governed by the resonance conditions of the frequencies. We also note that the softmax operator may act not as a damper but as a non-linear amplifier of resonance because small differences in logits are amplified due to the exponential nature.

Therefore, we employ the Hamiltonian formalism not as a global description of the flow, but as a local analysis of the resonance instability. Our argument is that the ``Attention Sink" represents a trivial fixed-point attractor caused by mode-locking (resonance). By satisfying the KAM non-resonance conditions via $\pi$-RoPE, Prism postpones this mode-locking, effectively maintaining a ``quasi-conservative" signal subspace where information propagates isentropically, postponing the collapse to the trivial attractor. 

Finally, we believe that de-noising and compression are the two fundamental imperatives in neural architecture design. Our results suggest that Prism, which is a white-box Transformer constructed from geometric principles, is not merely a modification of the standard Transformer, but a mathematically grounded alternative that achieves competitive performance while offering intrinsic interpretability. Our results indicate that interpretability and performance may be not necessarily trade-offs, but can be unified through principled geometric construction, where functional specialization emerges naturally from rigorous physical constraints.

Code will be available at \url{https://github.com/HuangDongchen/PRISM-Transformer}.

\section*{Acknowledgments}
This work was supported by the Postdoctoral Fellowship Program of
CPSF (Grant No. GZC20232943).

\bibliography{ref}
\bibliographystyle{icml2025}

\end{document}